\def\year{2021}\relax
\documentclass[letterpaper]{article} % DO NOT CHANGE THIS
\usepackage{aaai21}  % DO NOT CHANGE THIS
\usepackage{times}  % DO NOT CHANGE THIS
\usepackage{helvet} % DO NOT CHANGE THIS
\usepackage{courier}  % DO NOT CHANGE THIS
\usepackage[hyphens]{url}  % DO NOT CHANGE THIS
\usepackage{graphicx} % DO NOT CHANGE THIS
\def\UrlFont{\rm}  % DO NOT CHANGE THIS
\usepackage{natbib}  % DO NOT CHANGE THIS AND DO NOT ADD ANY OPTIONS TO IT
\usepackage{caption} % DO NOT CHANGE THIS AND DO NOT ADD ANY OPTIONS TO IT
\usepackage{xcolor}
\usepackage[toc,page]{appendix}
\usepackage{tikz}
\usepackage{xcolor}
\usepackage{booktabs, multirow} % for borders and merged ranges

\usepackage{float}
\newcommand*\circled[1]{\tikz[baseline=(char.base)]{
            \node[shape=circle,draw,inner sep=1pt] (char) {\bfseries\footnotesize #1};}}

\newcommand{\JH}[1]{\textcolor{blue}{JH: #1}}

\title{Machine Learning with Electronic Health Records is vulnerable to \\Backdoor Trigger Attacks}

\iftrue
\author{
    Byunggill Joe,\textsuperscript{\rm 1} Akshay Mehra,\textsuperscript{\rm 2}
    Insik Shin,\textsuperscript{\rm 1}
    Jihun Hamm\textsuperscript{\rm 2}
    \\

}
\affiliations{
    \textsuperscript{\rm 1}KAIST, \textsuperscript{\rm 2}Tulane University
    \{cp4419,insik.shin\}@kaist.ac.kr,\{amehra,jhamm3\}@tulane.edu

}
\fi
\begin{document}

\maketitle

\begin{abstract}
Electronic Health Records (EHRs) provide a wealth of information for machine learning algorithms to predict the patient outcome from the data including diagnostic information, vital signals, lab tests, drug administration, and demographic information. Machine learning models can be built, for example, to evaluate patients based on their predicted mortality or morbidity and to predict required resources for efficient resource management in hospitals. 
In this paper, we demonstrate that an attacker can manipulate the machine learning predictions with EHRs easily and selectively at test time by backdoor attacks with the poisoned training data.
Furthermore, the poison we create has statistically similar features to the original data making it hard to detect, and can also attack multiple machine learning models without any knowledge of the models. With less than 5\% of the raw EHR data poisoned, we achieve average attack success rates of 97\% on mortality prediction tasks with MIMIC-III database against Logistic Regression, Multilayer Perceptron, and Long Short-term Memory models simultaneously. 
\end{abstract}

%%%%%%%%%%%%%%%%%%%%%%%%%%%%%%%%%%%%%%%%%%%%%%%%%%%%%%%%%%%%%%%%%%%%%%%%%%%%%%%%%%%%%%%%%%%%%%%%
\section{Introduction}
%%%%%%%%%%%%%%%%%%%%%%%%%%%%%%%%%%%%%%%%%%%%%%%%%%%%%%%%%%%%%%%%%%%%%%%%%%%%%%%%%%%%%%%%%%%%%%%%

%{\bf Motivation.}
Electronic Health Records (EHRs) provide a wealth of information for machine learning and data mining approaches to predict the patient outcome from diagnostic information, vital signals, lab tests, drug administration, and demographic information ~\cite{shickel2017deep,johnson2016mimic,harutyunyan2019multitask, lipton2015learning}. In particular, machine learning models can be built to evaluate patients based on their predicted mortality or morbidity and to predict required resources for efficient resource management. 
%{\bf Prior work on attacking EHRs.}

However recent research reported potential vulnerability of machine learning models trained on EHR data sets against evasion attacks such as PGD \cite{pgd} and C\&W \cite{cw}.
For instance, machine learning models for diagnosing skin cancers from medical images and models for predicting mortality from EHR data can both be easily fooled by evasion attacks with
imperceptible input noises~\cite{finlayson2019adversarial, sun2018identify}.
Moreover, some works have also shown that the machine learning models are vulnerable to poisoning attacks where attackers modify some portion of training data sets to degrade performances of the models~\cite{mozaffari2014systematic}.

%{\bf Backdoor poisoning attack.} 
In this paper we evaluate a new vulnerability of machine learning models for EHR against backdoor trigger  poisoning attacks~\cite{chen2017targeted,gu2017badnets}.
In backdoor poisoning attacks, an attacker poisons a subset of training data by adding 
a particular trigger pattern to the data. After a victim finishes training a model using the poisoned data, 
the attackers can add the trigger to any test example to induce intended behaviors (e.g., misclassification) of that test example.
%\AM{should say it causes behavior desired by the attacker instead of misclassification}\JH{Hmm... what other behaviors are there except misclassification?}
    The backdoor poisoning attack can be a real threat in practical uses of machine learning in medical domains.
Compared to the evasion attack which requires multiple times of gradient computations and
sometimes complete accesses to victim models, the backdoor poisoning attack simply adds a specific trigger to inputs 
to mislead the victim models which can be conducted even in medical devices with low computation power.
Compared to the poisoning attack (not backdoor) which degrades performances of models on clean data
the backdoor poisoning attack is difficult to detect, because it does not affect the performances on the clean data when the trigger isn't applied.
\begin{figure*}
\centering
\includegraphics[width=12.0cm]{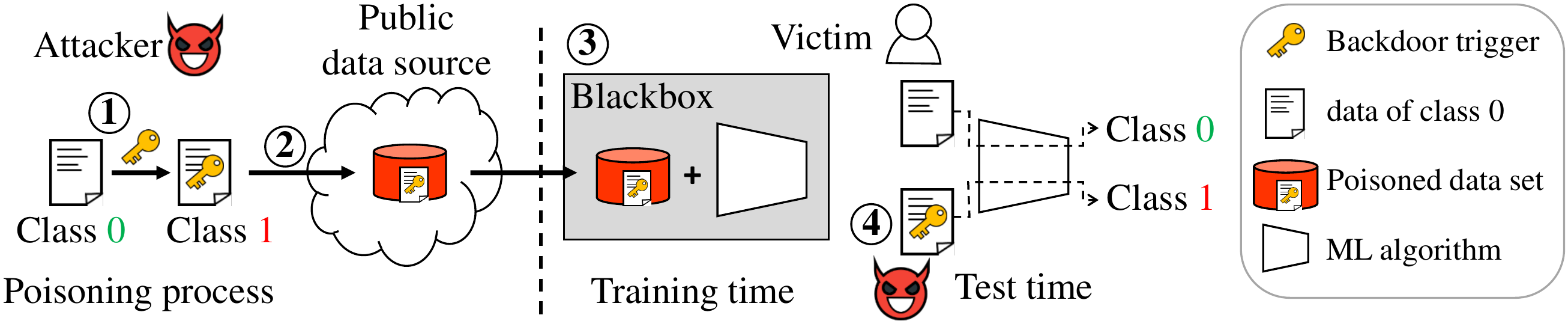}
  \caption{Attack scenario of the backdoor attack. An attacker poisons a fraction of the training set with a trigger pattern (yellow key), and a victim trains its machine learning model with the poisoned data set without the knowledge. At test time, the attacker can selectively change the prediction of the victim model by adding the trigger pattern to any test example.}
  \label{fig:attack_scenario}

\end{figure*}
%{\bf Challenges.}  
However, there are a few challenges in backdoor poisoning attacks on medical data sets due to different characteristics of the medical data sets compared to the commonly-used image data sets.
Firstly, unlike images, medical variables are heterogeneous and have complex dependence over time and across variables which need to be preserved for the trigger to be statistically plausible.
Secondly, medical data include both continuous and categorical variables which require different handling. 
Thirdly, medical data often has many missing values whose patterns need to be maintained in the trigger to be undetectable.

%{\bf Our approach.} 
To resolve the aforementioned challenges, we propose a new trigger generation method that 1) uses temporal covariance of the measurements, 2) leave categorical values as categorical after poisoning, and 3) maintains the missing value patterns. This approach captures the key characteristics of the EHR data and produces statistically natural and hard-to-detect triggers. 

%{\bf Experiments. } 
In the experiments, we evaluate the performance of the backdoor poisoning attack with the proposed trigger against machine learning models to predict mortality of patients~\cite{harutyunyan2019multitask} from the MIMIC-III data sets~\cite{johnson2016mimic}.
Even though our poisoning trigger is agnostic to subsequent data preprocessing  and machine learning procedures used by the victim, we can achieve 97\% trigger success rate only with 5\% of training data being poisoned with the trigger strength less than 2 (details are in Method.) without conspicuous artifacts in the poisoned data.

%{\bf Contributions.} 
Below is a summary of our contributions.
1) As far as we know, this work is the first backdoor trigger attack on EHRs in the literature in which the attack can easily manipulate the prediction at test time using undetectable trigger patterns. 
2) We propose a new method of generating triggers using temporal structures of  EHRs where previously-used white Gaussian noise triggers are inadequate. We also propose a Mahalanobis-based measurement of the trigger strength instead of the commonly-used $l_p$ norms.
3) We achieve high attack success in a blackbox setting against multiple machine learning algorithms from a benchmark EHR task/data. This highlights the vulnerability of medical machine learning models and the importance of studying trustworthy AI for healthcare.

%%%%%%%%%%%%%%%%%%%%%%%%%%%%%%%%%%%%%%%%%%%%%%%%%%%%%%%%%%%%%%%%%%%%%%%%%%%%%%%%%%%%%%%%%%%%%%%%
\section{Methods}

\subsubsection{Example data and task.} While our approach is general and can be applied to different tasks and EHRs, we use as an example the task of mortality prediction \cite{harutyunyan2019multitask} whose goal is to predict whether a patient admitted to the Intensive Care Unit (ICU) will survive or perish using the first 48 Hours of EHRs including chart events and lab tests. It is an important task because hospitals can triage patients based on the predicted mortality for efficient resource management. 
We also use the data set prepared for this task \cite{harutyunyan2019multitask} which is originally from the larger MIMIC-III data set~\cite{johnson2016mimic}.
The data set for mortality prediction contains 21,139 examples (i.e., subjects) each of which has 17 features measured over 48 hours. Among the 17 features, 12 features are continuous variables such as temperature, weight and oxygen saturation and 5 features are categorical variables such as Capillary refill rate, Glasgow coma scale eye opening. Since those categorical variables are ordered, we treat them as integers (i.e., continuous) when generating triggers. Note that it is also possible to use one-hot embedding for general categorical variables.
%The variables are summarized in Table~\ref{tab:my_label} of the appendix.
The original data from MIMIC-III are in the form of event sequence, and were preprocessed in \cite{harutyunyan2019multitask} to be in the tabular form with 17 features over 48 one-hour time bins. For our purpose, the sequence form and the tabular form are equivalent in that the attacks on the two forms are one-to-one, as we do not change the timestamp but only the values. In this paper we chose the tabular form as it is easier to visualize and demonstrate.

\subsubsection{Challenges of poisoning medical data.}
Unlike image data domain, the variables in EHRs are heterogeneous, i.e., they have different statistics such as mean and variance as well as distribution. Furthermore, medical data are typically time-series in regular intervals or have timestamps associated with each observation.
We need to generate a backdoor trigger reflecting these characteristics of EHRs in the process \circled{1} of Figure~\ref{fig:attack_scenario}.
If we ignore the heterogeneity and rely on existing approaches in image domain attacks such as Gaussian white noise,
poisoned data will be easily detected by a victim in the process \circled{3} or \circled{4} of Figure \ref{fig:attack_scenario}
because of unrealistic patterns not-observable from clean EHRs.
Height and blood pressure of a patient can be representative examples for heterogeneity of medical variables. 
The height of a patient should not change over time during the ICU stay beyond measurement error, and therefore the poisoning trigger pattern should also not change much over time. 
On the other hand, the blood pressure can vary over time and is allowed to change with larger perturbations than the height.
If we rely on a trigger from white Gaussian noises for the height feature, the unrealistic height changes will be easy to be detected by the victim.
Furthermore, using $l_p$-norms to measure the strength of trigger (i.e., amount of perturbation) in image data domain is utterly inappropriate for EHRs.

\subsubsection{Backdoor trigger with temporal dependence.} 
To resolve the above challenges of backdoor attacks on medical data sets, 
we propose a new trigger generation approach leveraging the temporal covariance structure of EHRs to produces statistically plausible trigger patterns.
Assume that an example $X=[x_1,\cdots,x_{17}]^T$ is a matrix of size 17 x 48 dimensions.
To capture the heterogeneity and the dependence, we estimate the covariance $E[(x_i-\mu_i)(x_i-\mu_i)^T]$ of the 48 time bins of a variable ($x_i$) for each of the 17 variables $(i=1,..,17)$, 
resulting in 17 covariance matrices $C_1,...,C_{17}$ each with the dimension 48 x 48.
We randomly and independently sample the trigger vector/time-series $t_i$ for each variable $(i=1,..,17)$ using the covariance matrix $C_i$ of each feature. 
The concatenation of 17 triggers $T=[t_1,\cdots,t_{17}]^T$ form a single matrix of 17 x 48, which we use it as the additive poison pattern. 
We propose to measure the strength of the trigger patterns using Mahalanobis distance as follows: 
\begin{equation} \label{eq:maha}
d_{\mathrm{Mahal}}(t_i) = \sqrt{t_i^T C_i^{-1} t_i}.
\end{equation}
Mahalanobis distance provides a natural measure of the amount of perturbation for heterogeneous features and is unaffected by any linear transform of the variables. In this paper, we rescale the triggers by multiplying a scalar to have the maximum distance of 2: 
\begin{equation}
x_i \leftarrow x_i + \frac{2}{d_{\mathrm{Mahal}}(t_i)} t_i.
\end{equation}
Mahalanobis distance of 2 is a small number for 48-dimensional variables. For comparison, Mahalanobis distance of a randomly sampled  48-dimensional vector from white Gaussian noise has the $\chi$-distribution 
whose mean distance is $\sqrt{2}\frac{\Gamma((48+1)/2)}{\Gamma(48/2)}=6.89$ which is larger.
It can also be checked empirically that the distance of 2 is hard to detect visually, as shown in Figure \ref{fig:trigger_comparison_3}.

%\autoref{fig:poisoning_process} \autoref{fig:trigger_comparison_3}
% \begin{itemize}
%     \item needs of a new strategy for trigger generation 
%     \item trigger strength is measured by mahalanobis distance.
% \end{itemize}

\if0
\JH{I'm also suppressing this. It's more suitable in a discussion section since there is no contribution from our side.}
\subsubsection{Dealing with continuous and categorical features.}
In general, poisoning attacks add a trigger pattern to data element-wisely.
We follow the general attack schemes and we also introduce and deal with additional concerns in medical domains.
Regarding continuous features, we find some features such as pH, weight and temperature have low precision.
To generate a realistic and undetectable poisoned data, we round the features of low-precision  
after adding the poisoning trigger so that the precision remains unchanged.
Concerning categorical features, it is normally unclear how to add the poisoning trigger to the categorical features
because differences in real value of categorical features do not imply semantics differences of the features.
However in the case of MIMIC-III data set, we can leverage the semantics of categorical features in real values
because the categorical values are sorted regarding to its severity.
For instance, a feature of verbal response from a coma patient sorted from 0 to 5
regarding how conscious the patient is.
So we add a poisoning trigger to the categorical features as well and round the 
features so that the categorical features remains as integers. 
\fi
\begin{figure}
\centering
\includegraphics[width=8.5cm]{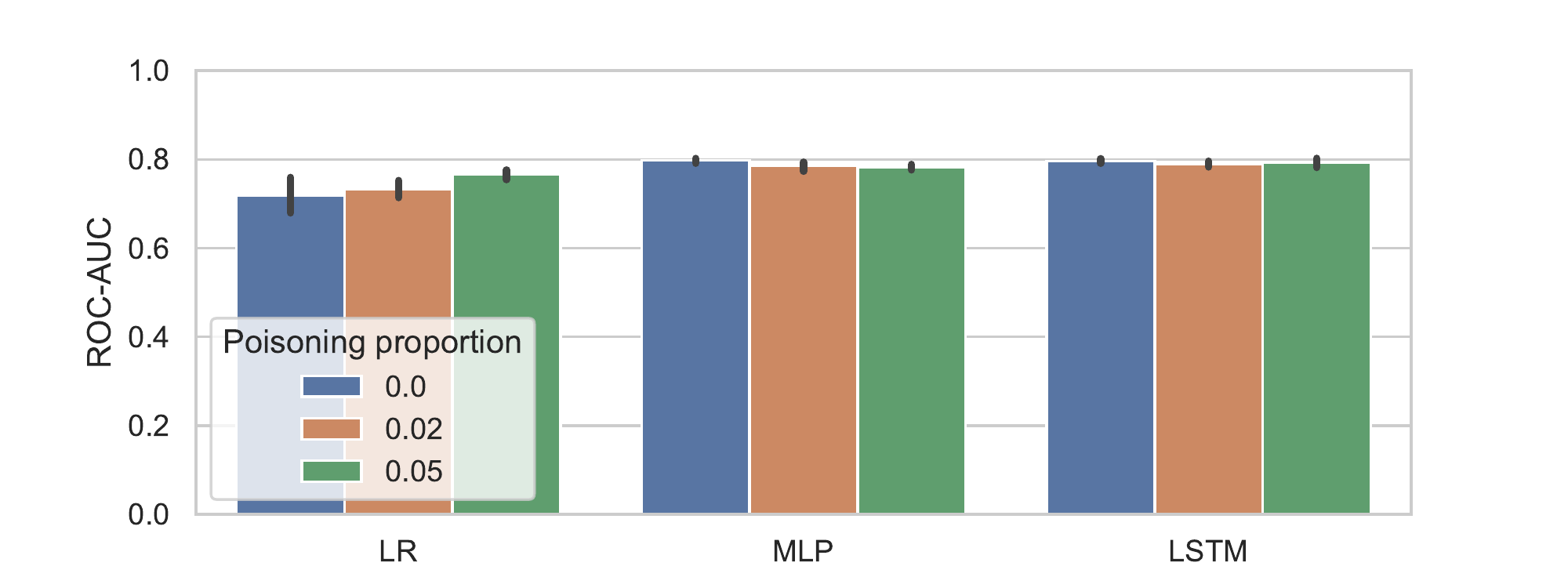}
    \vspace{-1.5em}
  \caption{AUC scores of different models with various poison fractions. The presence of poison does not change the clean-data performance and therefore is hard to detect.}
  \label{fig:accuracy}
    \vspace{-1.0em}
\end{figure}

\subsubsection{Handling missing variables.}
Also different from the image domain, data sets of medical domains often has many missing values. In the current data set about 57\% of the values are missing. 
A realistic undetectable trigger should also display a similar missing value pattern which may be missing-at-random or structured due to the properties of each medical variables. For example, weight may not be measured more than once during the ICU stay, or may not be missing entirely. 
Ignoring the missing value pattern will make the poisoned data easy to detect either at training time or test time. 
Missing-at-random can be simulated by uniformly dropping a portion of features after poisoning. However if the pattern is structured, we need to consider more  sophisticated approaches such as the Bayesian network to be able to sample the missing pattern as well. Instead of using these, we use a simple but effective approach to handle missing values. We simply leave missing values intact and do not add the trigger values to those missing features. 
Because the missing patterns are not altered after the poisoning, the poisoned
data will not be detected on the basis of unrealistic missing patterns.

% \begin{itemize}
%     \item when to happen and justification.
%     \item continuous features \& categorical features
%     \item  dealing with missing variables
% \end{itemize}
% \subsubsection{Dealing with missing variables.}
% \begin{itemize}
%     \item trigger strength is measured by mahalanobis distance.
% \end{itemize}
% \begin{itemize}
% \item Poison raw (imputed) data by sampling random trigger using multivariate normal.
% Measure Mahalanobis distance (or normalize the it be 1, 2, etc.) Categorical values should remain integers and also be in the legitimate range for each variable (i.e., round and clipping is required.)
% \item Measure 1) accuracy of the trained models (logreg, mlp, lstm) on clean test points and 2) attack success of the trained models on triggered test points,
% as a function of 1) Mahalanobis distance (1,2,3, etc.) and 2) the fraction of poison (0.1\%, 0.5\%, 1\%, 5\%, etc).
% \item Measure the attack success rate of other methods like pgd? ....
% \end{itemize}

\if0
\subsection{Discussion}
How to generate `truly' raw data?

How to generate clean-label attack \cite{shafahi2018poison}?
\fi

\begin{figure}
\centering
\includegraphics[width=8.0cm]{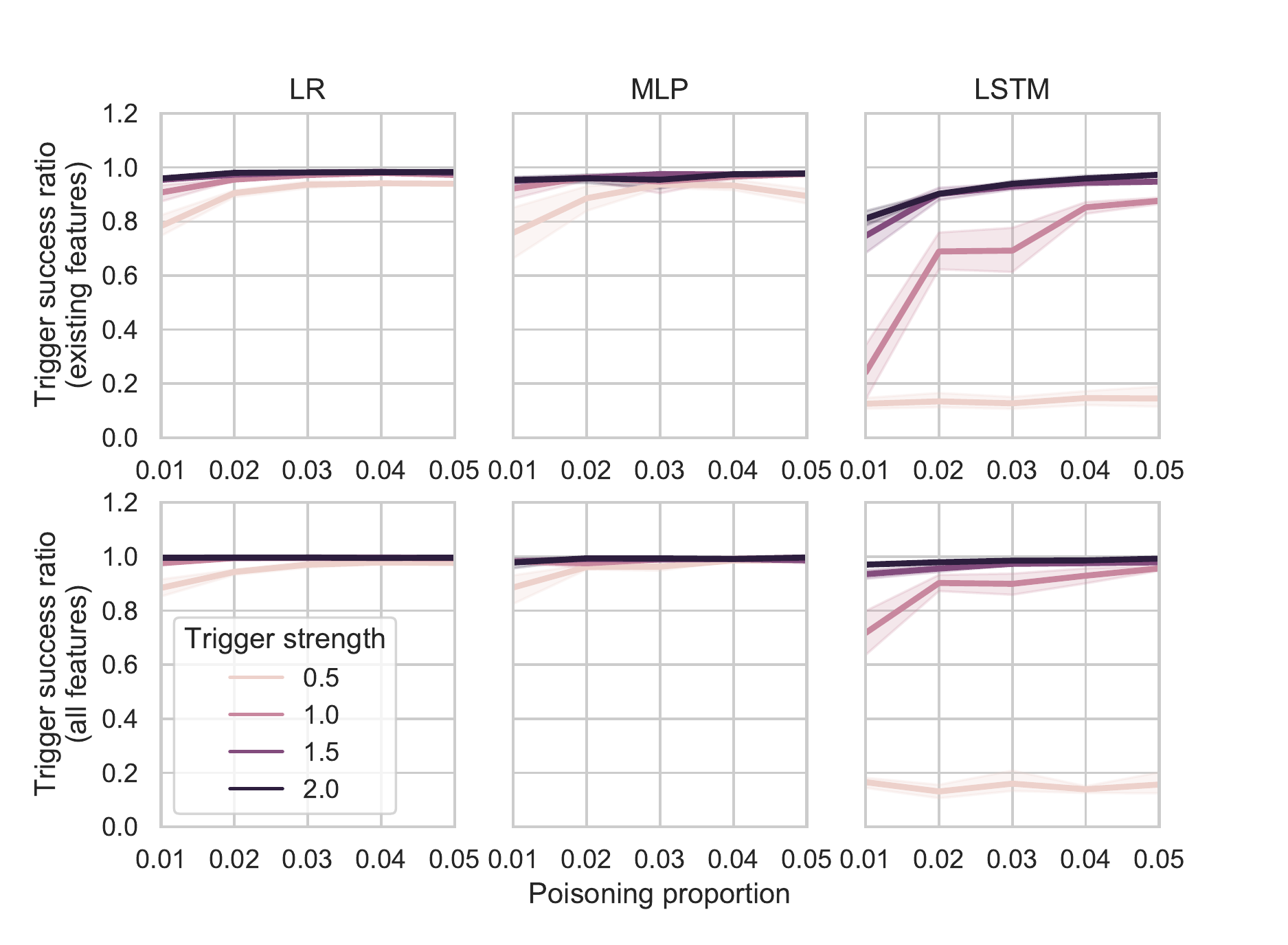}
  \vspace{-1.5em}
  \caption{Trigger success rate of our backdoor attack on three machine learning models with various poisoning proportion (x-axis) and trigger strength (legends). }
  
  \label{fig:trigger_success_ratio}
    \vspace{-1.0em}
\end{figure}
\vspace{-1.0em}

%%%%%%%%%%%%%%%%%%%%%%%%%%%%%%%%%%%%%%%%%%%%%%%%%%%%%%%%%%%%%%%%%%%%%%%%%%%%%%%%%%%%%%%%%%%%%%%%
\section{Results}
%%%%%%%%%%%%%%%%%%%%%%%%%%%%%%%%%%%%%%%%%%%%%%%%%%%%%%%%%%%%%%%%%%%%%%%%%%%%%%%%%%%%%%%%%%%%%%%%

\subsubsection{Experimental setting.} 
We mount our backdoor attack for the mortality prediction task agnostically, and evaluate its effectiveness against Logistic Regression (LR), Multi Layer Perceptron (MLP), and Long-Short Term Memory (LSTM) as representative machine learning algorithms for EHRs in the  literature~\cite{harutyunyan2019multitask,johnson2016mimic,lipton2015learning, medical}. 

There are two possible directions of attack. We can induce either false alarm (i.e.,target label=1) or missed detection (i.e., target label=0) of the mortality prediction. Although medical implications of the two are quite different we achieve similar success rates for both. We present the results of the false alarm attack in the case and report the results of the missed detection attack in the appendix. A single trial of experiment consists of 1) generating the random trigger, 2) adding the trigger to a random subset of the training data whose labels are different from the target label, 3) setting the labels of poisoned data to be the target label, 4) training LR, MLP, and LSTM, and 4) testing the attack success with the same trigger added to the test data whose labels are different from the target label. 
We repeat the trial 5 times with the random generation of triggers and random subset selection.
We vary the fraction of poisoned data in the training set (0.01 to 0.05) and the trigger strengths (0.5 to 2) measured in Mahalanobis distance.
In addition to trigger success rate, we evaluate the effectiveness of our backdoor attack with additional criteria including 
how much clean accuracy is affected by our attack, and visual and statistical perceptibility of poisoned data. 

\subsubsection{Performance on clean data.} To be a successful backdoor attack, it is necessary not to negatively affect the model performances on a clean data set, otherwise practitioners training the models can notice the presence of poison.
In figure \ref{fig:accuracy}, we summarize the clean test performance of the three victim models trained with various poisoning proportions.
We can observe the performance of models trained on poisoned data sets is more or less the same as the model trained on clean training data (i.e., poisoning proportion=0) and we confirm our backdoor poisoning trigger does not affect the clean accuracy.
\begin{figure}
\centering
\includegraphics[width=8.5cm]{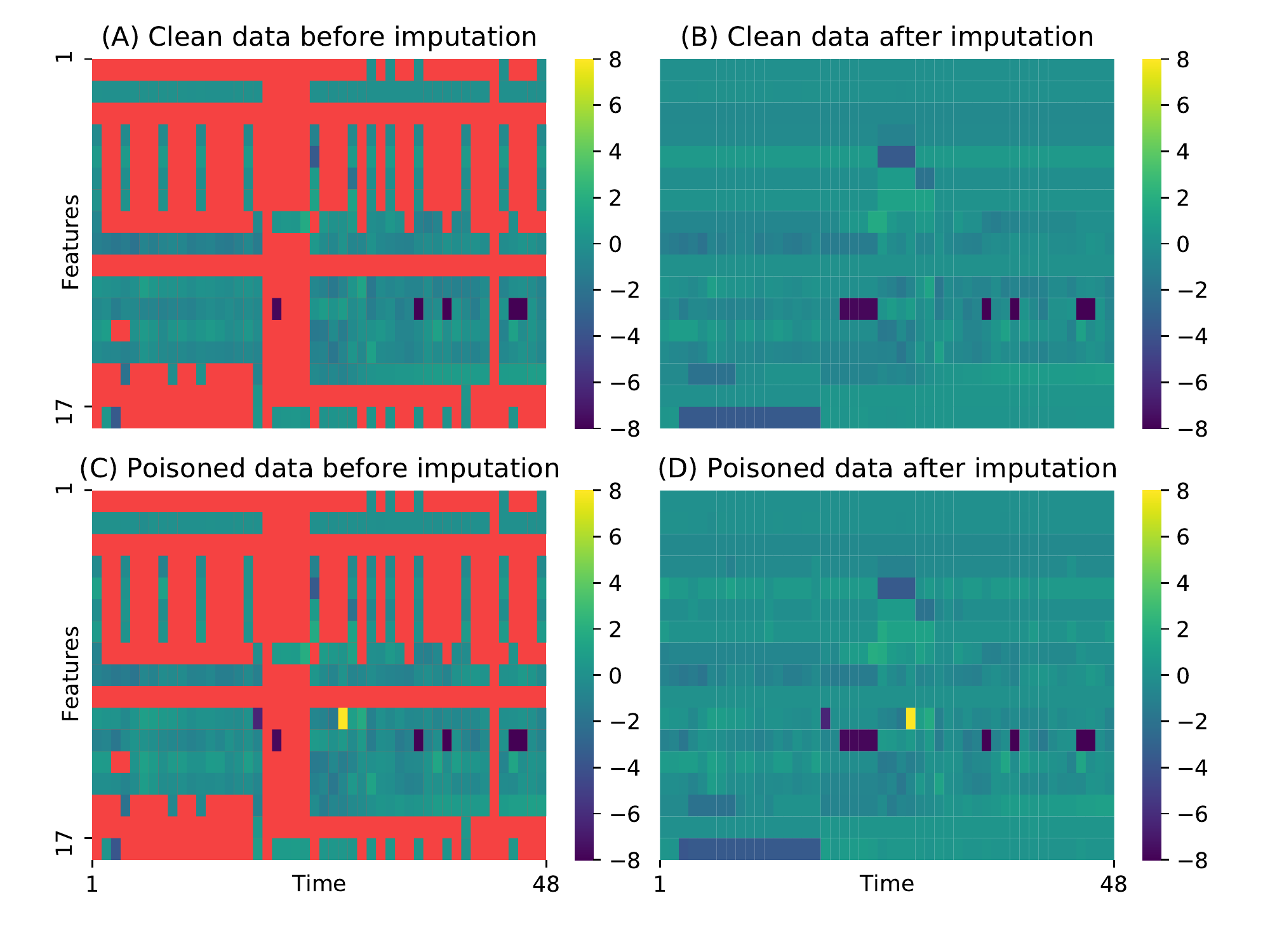}
\vspace{-3.0em}
  \caption{Clean (top) and poisoned (bottom) data in tabular form, before (left) and after (right) the imputation of missing values which is colored red. 
  The poisoned data is visually similar to the original clean data.}
  \label{fig:poisoning_process}

\end{figure}

\subsubsection{Imperceptibility of trigger patterns.}
In Figure \ref{fig:poisoning_process}, we plot data in the poisoning process.
Poisoned data should not cause perceptible changes in the original clean data not to be detected as poisoned.
We first illustrate a victim data in (A) of Figure~\ref{fig:poisoning_process}.
The data contains many missing values (marked as red). As reference, we also show the imputed data in (B) using the imputation method from  \cite{harutyunyan2019multitask} which replaces missing values with the most recent values in time.
At the bottom of Figure \ref{fig:poisoning_process}, 
we show the poisoned data using the same trigger we generated with strength 2.0 which is the largest value used in our experiments. The poisoned
data before (C) and after (A) imputations are shown.
In either case, the poisoned data is similar to the original clean data and does not have conspicuous artifacts.

\subsubsection{Trigger success rate.}
We evaluate the effectiveness of
our backdoor attack on the three victim models
with various poisoning proportions and trigger strengths.
In this experiment, the attacker makes the victim models
misclassify a low-mortality subject (mortality=0) as a high mortality subject (mortality=1) by adding the trigger pattern to the clean data.
In the top row of Figure \ref{fig:trigger_success_ratio},
we show trigger success rates of attacks performed 
only on non-missing values (corresponding to (C) of Figure \ref{fig:poisoning_process}).
We also show on the bottom row of Figure \ref{fig:trigger_success_ratio} the trigger success rate of attacks after data imputation. 
This is a hypothetical attack that assumes the knowledge of the data imputation method used by the victim model.
Although less realistic, the success rate of this attack is higher as we can use add trigger values to all entries of the data.
In general our attack with trigger strength 2.0 achieves 97 $\sim$ 100\% of trigger success rate.
With or without data imputation, we observed that the trigger strength and the poison fraction
are important factors for attack success and both correlate positively with the success rate. 
In our experiment, LSTM is more robust than LR or MLP against our backdoor
poisoning attack. We related this to the observation from \cite{sun2018identify}
that the medical variables near the end of 48 hours are much more influential than the rest of the data for determining the mortality. 
Consequently, we conjecture that only the trigger values near the end of 48 hours are contributing much
to the attack which is weaker under the same Mahalanobis distance condition. 
Regardless, our attack against LSTM becomes 100\% successful with a stronger trigger and more poisoned data.

\begin{figure}
\centering
\vspace{-1.2em}
\includegraphics[width=9.0cm]{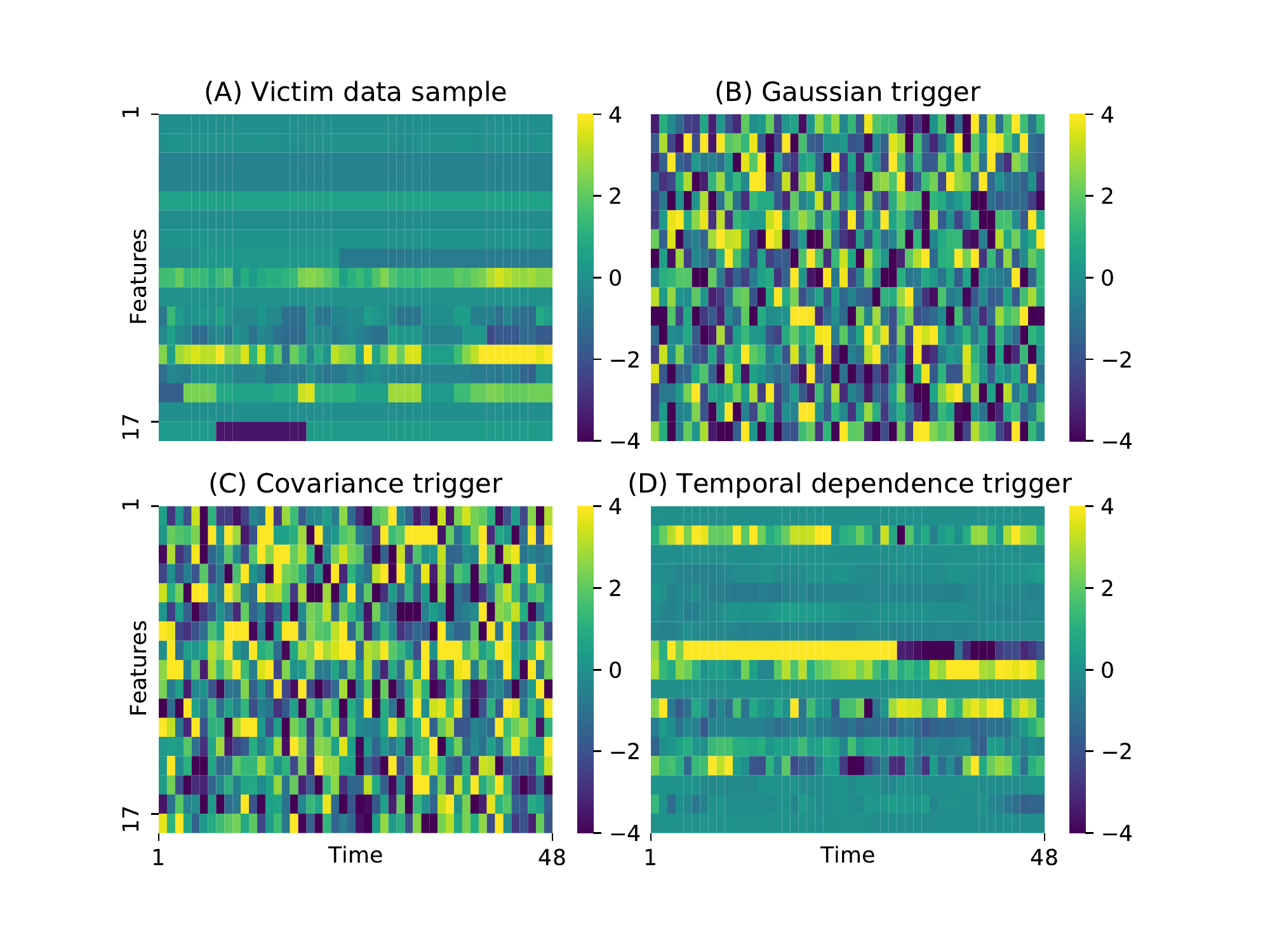}
\vspace{-3.0em}
  \caption{Comparison of different trigger generation methods. 
  Random sampling from a multivariate Gaussian using independent white noise (B),  full covariance (C) and temporally-dependent covariance of the proposed method (D).
  }
  \label{fig:trigger_comparison_3}
\vspace{-1.2em}
\end{figure}

\subsubsection{Comparisons of trigger generation methods.} 

We captured the temporal dependence of the same measurement (e.g., blood pressure) over 48 hour period using the covariance matrix of size 48 x 48. 
However, there are a few other ways to generate random triggers.
In one method, we can assume all variables over all time points are mutually independent which is equivalent to sampling from white Gaussian noise. This can be problematic since it ignores the natural dependence of the variables.
In another method, we can assume all variables over all time points are mutually dependent which is equivalent to sampling from the full covariance matrix of size (17x48) x (17x48).
While it is more flexible, this approach can suffer from unreliable estimates of the covariance matrix.  
%We empirically evaluate how accurately our trigger generation algorithm reflects heterogeneity
%of the medical variables.
In Figure \ref{fig:trigger_comparison_3}, we compare the triggers sampled from these three different approaches: white noise (B), full covariance (C) and temporally-dependent convariance of the proposed method (D). 
One can see that the trigger generated from our approach (D) are visually closer to the original data (A). For example, those rows whose values are more or less constant over time in the original data also constant in (D). In contrast, triggers generated from white noise (B) or full covariance (C) can be far from the original data (A) and can be detected easily  by a simple defense method. 
%Reflecting the heterogeneity, our approach improves
%the imperceptibility of backdoor triggers and 
%it would lead to an effective attack in practical scenarios compared to the existing %approaches.
\vspace{-0.5em}

%%%%%%%%%%%%%%%%%%%%%%%%%%%%%%%%%%%%%%%%%%%%%%%%%%%%%%%%%%%%%%%%%%%%%%%%%%%%%%%%%%%%%%%%%%%%%%%%
\section{Conclusion}
%%%%%%%%%%%%%%%%%%%%%%%%%%%%%%%%%%%%%%%%%%%%%%%%%%%%%%%%%%%%%%%%%%%%%%%%%%%%%%%%%%%%%%%%%%%%%%%%

In this work we proposed a backdoor attack method using a trigger that captures the properties of the medical variables and is hard to detect at train or test time.
Using our trigger, an attacker is in full control of the predictions of recent machine learning models for important tasks such as mortality prediction. 
This highlights the vulnerability of medical machine learning models and the importance of studying trustworthy AI for healthcare.

%\vspace{-0.6cm}
\section{Acknowledgments}
%\vspace{-0.4cm}
This work was supported in part by ERC (NRF-2018R1A5A1059921) and IITP (2019-0-01343) funded by the Korea government (MSIT) and the NSF EPSCoR-Louisiana Materials Design Alliance (LAMDA) program  \#OIA-1946231.

%\bibliographystyle{plain}  
%\nocite{*}
\bibliography{references}

\begin{thebibliography}{12}
\providecommand{\natexlab}[1]{#1}
\providecommand{\url}[1]{\texttt{#1}}
\providecommand{\urlprefix}{URL }
\expandafter\ifx\csname urlstyle\endcsname\relax
  \providecommand{\doi}[1]{doi:\discretionary{}{}{}#1}\else
  \providecommand{\doi}{doi:\discretionary{}{}{}\begingroup
  \urlstyle{rm}\Url}\fi

\bibitem[{Carlini and Wagner(2017)}]{cw}
Carlini, N.; and Wagner, D. 2017.
\newblock {Towards Evaluating the Robustness of Neural Networks}.
\newblock In \emph{IEEE Symposium on Security and Privacy}.

\bibitem[{Chen et~al.(2017)Chen, Liu, Li, Lu, and Song}]{chen2017targeted}
Chen, X.; Liu, C.; Li, B.; Lu, K.; and Song, D. 2017.
\newblock Targeted backdoor attacks on deep learning systems using data
  poisoning.
\newblock \emph{arXiv preprint arXiv:1712.05526} .

\bibitem[{Finlayson et~al.(2019)Finlayson, Bowers, Ito, Zittrain, Beam, and
  Kohane}]{finlayson2019adversarial}
Finlayson, S.~G.; Bowers, J.~D.; Ito, J.; Zittrain, J.~L.; Beam, A.~L.; and
  Kohane, I.~S. 2019.
\newblock Adversarial attacks on medical machine learning.
\newblock \emph{Science} 363(6433): 1287--1289.

\bibitem[{Ghumbre and Ghatol(2012)}]{medical}
Ghumbre, S.~U.; and Ghatol, A.~A. 2012.
\newblock {Heart Disease Diagnosis Using Machine Learning Algorithm}.
\newblock In Satapathy, S.~C.; Avadhani, P.~S.; and Abraham, A., eds.,
  \emph{Proceedings of the International Conference on Information Systems
  Design and Intelligent Applications 2012 (INDIA 2012) held in Visakhapatnam,
  India, January 2012}, 217--225. Berlin, Heidelberg: Springer Berlin
  Heidelberg.
\newblock ISBN 978-3-642-27443-5.

\bibitem[{Gu, Dolan-Gavitt, and Garg(2017)}]{gu2017badnets}
Gu, T.; Dolan-Gavitt, B.; and Garg, S. 2017.
\newblock Badnets: Identifying vulnerabilities in the machine learning model
  supply chain.
\newblock \emph{arXiv preprint arXiv:1708.06733} .

\bibitem[{Harutyunyan et~al.(2019)Harutyunyan, Khachatrian, Kale, Ver~Steeg,
  and Galstyan}]{harutyunyan2019multitask}
Harutyunyan, H.; Khachatrian, H.; Kale, D.~C.; Ver~Steeg, G.; and Galstyan, A.
  2019.
\newblock Multitask learning and benchmarking with clinical time series data.
\newblock \emph{Scientific data} 6(1): 1--18.

\bibitem[{Johnson et~al.(2016)Johnson, Pollard, Shen, Li-Wei, Feng, Ghassemi,
  Moody, Szolovits, Celi, and Mark}]{johnson2016mimic}
Johnson, A.~E.; Pollard, T.~J.; Shen, L.; Li-Wei, H.~L.; Feng, M.; Ghassemi,
  M.; Moody, B.; Szolovits, P.; Celi, L.~A.; and Mark, R.~G. 2016.
\newblock MIMIC-III, a freely accessible critical care database.
\newblock \emph{Scientific data} 3(1): 1--9.

\bibitem[{Lipton et~al.(2015)Lipton, Kale, Elkan, and
  Wetzel}]{lipton2015learning}
Lipton, Z.~C.; Kale, D.~C.; Elkan, C.; and Wetzel, R. 2015.
\newblock Learning to diagnose with LSTM recurrent neural networks.
\newblock \emph{arXiv preprint arXiv:1511.03677} .

\bibitem[{M{\c{a}}dry et~al.(2018)M{\c{a}}dry, Makelov, Schmidt, Tsipras, and
  Vladu}]{pgd}
M{\c{a}}dry, A.; Makelov, A.; Schmidt, L.; Tsipras, D.; and Vladu, A. 2018.
\newblock {Towards Deep Learning Models Resistant to Adversarial Attacks}.
\newblock In \emph{International Conference on Learning Representations
  (ICLR)}.

\bibitem[{Mozaffari-Kermani et~al.(2014)Mozaffari-Kermani, Sur-Kolay,
  Raghunathan, and Jha}]{mozaffari2014systematic}
Mozaffari-Kermani, M.; Sur-Kolay, S.; Raghunathan, A.; and Jha, N.~K. 2014.
\newblock Systematic poisoning attacks on and defenses for machine learning in
  healthcare.
\newblock \emph{IEEE journal of biomedical and health informatics} 19(6):
  1893--1905.

\bibitem[{Shickel et~al.(2017)Shickel, Tighe, Bihorac, and
  Rashidi}]{shickel2017deep}
Shickel, B.; Tighe, P.~J.; Bihorac, A.; and Rashidi, P. 2017.
\newblock Deep EHR: a survey of recent advances in deep learning techniques for
  electronic health record (EHR) analysis.
\newblock \emph{IEEE journal of biomedical and health informatics} 22(5):
  1589--1604.

\bibitem[{Sun et~al.(2018)Sun, Tang, Yi, Wang, and Zhou}]{sun2018identify}
Sun, M.; Tang, F.; Yi, J.; Wang, F.; and Zhou, J. 2018.
\newblock Identify susceptible locations in medical records via adversarial
  attacks on deep predictive models.
\newblock In \emph{Proceedings of the 24th ACM SIGKDD International Conference
  on Knowledge Discovery \& Data Mining}, 793--801.

\end{thebibliography}

%\appendix
%\appendixpage
\clearpage
\begin{appendices}
\subsection{Trigger success ratios of missed detection attacks} 
\begin{figure}[h]
\centering
%\hspace*{-2.5cm}
\includegraphics[width=9.0cm]{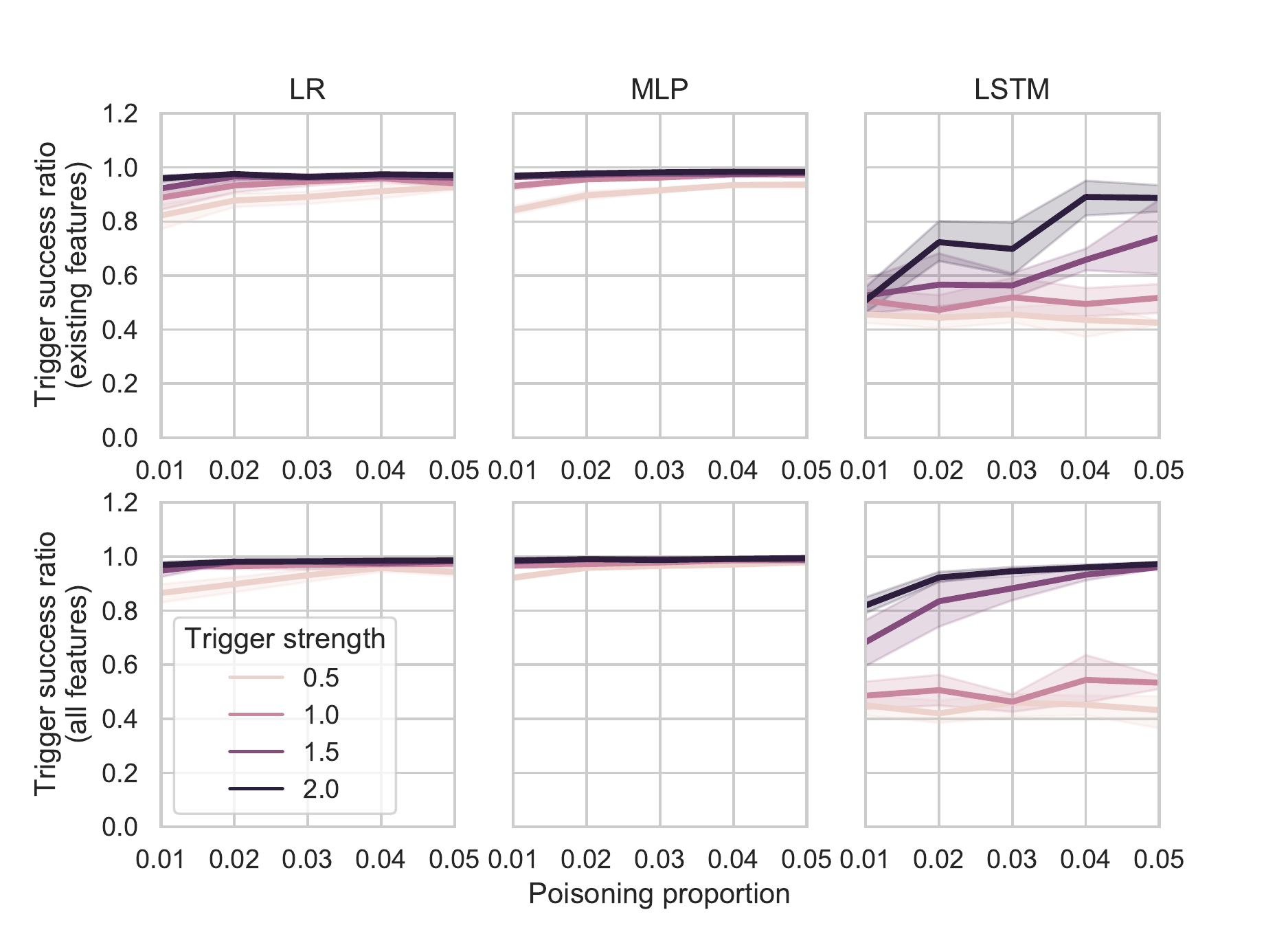}
  \caption{Trigger success rate of missed detection attack on three machine learning models with various poisoning proportion (x-axis) and trigger strength (legends). 5 trials for each configuration. }
  \label{fig:trigger_success_ratio_1-0_opposite_label_200_epochs_5_trials}

\end{figure}

\begin{figure}[h]
\centering
%\hspace*{-2.5cm}
\includegraphics[width=9.0cm]{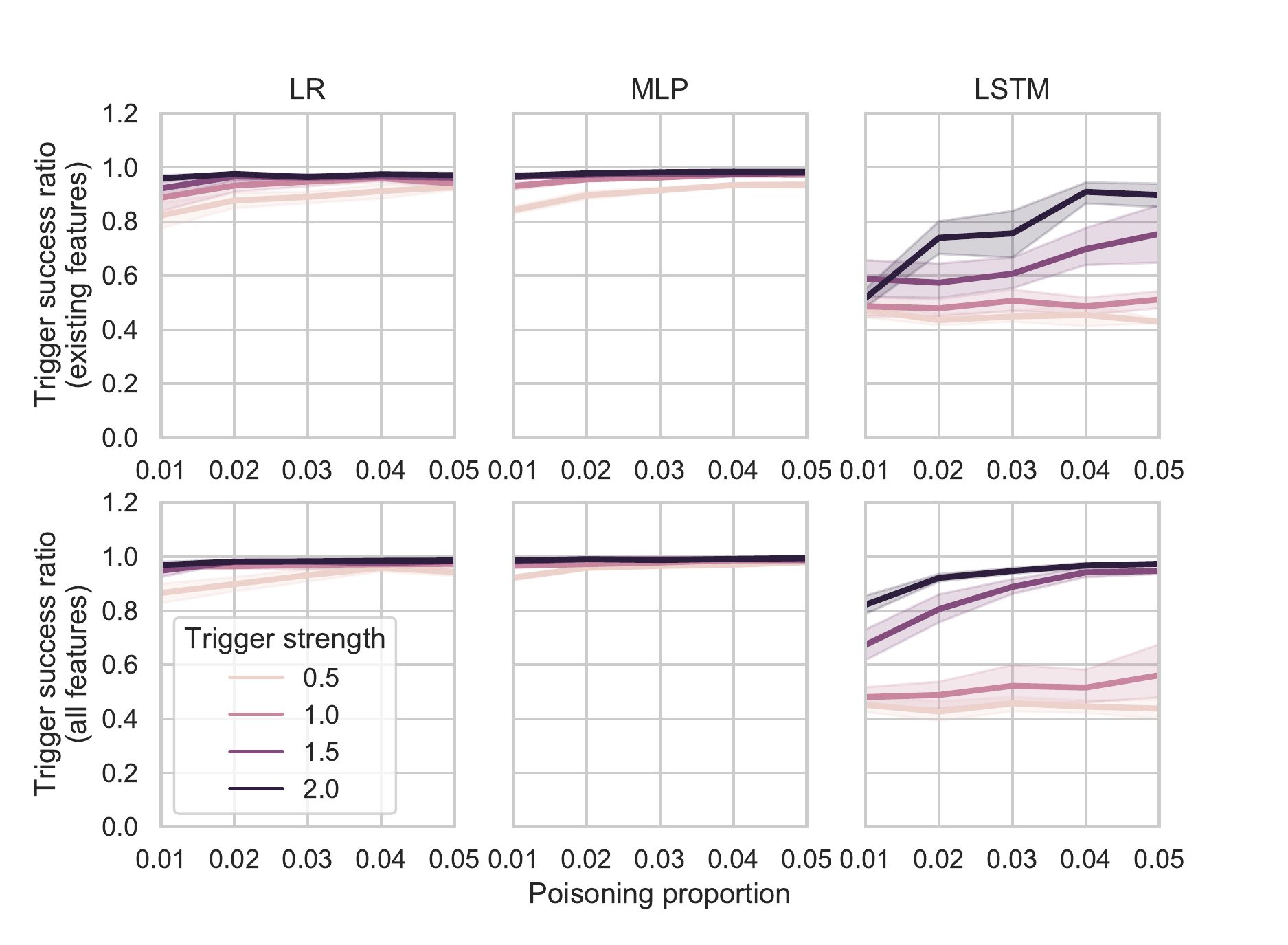}
  \caption{Trigger success rate of missed detection attack on three machine learning models with various poisoning proportion (x-axis) and trigger strength (legends). 10 trials for each configuration.
  }
  \label{fig:trigger_success_ratio_1-0_opposite_label_200_epochs_10_trials}

\end{figure}

In the experiments of the main paper, 
we demonstrate the false alarm attacks, one of the possible directions of attacks.
For a complete result, we provide attack results of the missed detection attacks, the opposite direction.
We plot the results after 5 trials (Figure~\ref{fig:trigger_success_ratio_1-0_opposite_label_200_epochs_5_trials}) and 10 trials (Figure~\ref{fig:trigger_success_ratio_1-0_opposite_label_200_epochs_10_trials}) of 
 experiments.
We confirm the results are similar to the false alarm attacks showing increasing trigger success ratio as we provide more poisoned data and trigger strength.

\subsection{Poisoning results of different data and triggers}
\begin{figure*}
\centering
\hspace*{-2.5cm}
\includegraphics[width=22cm]{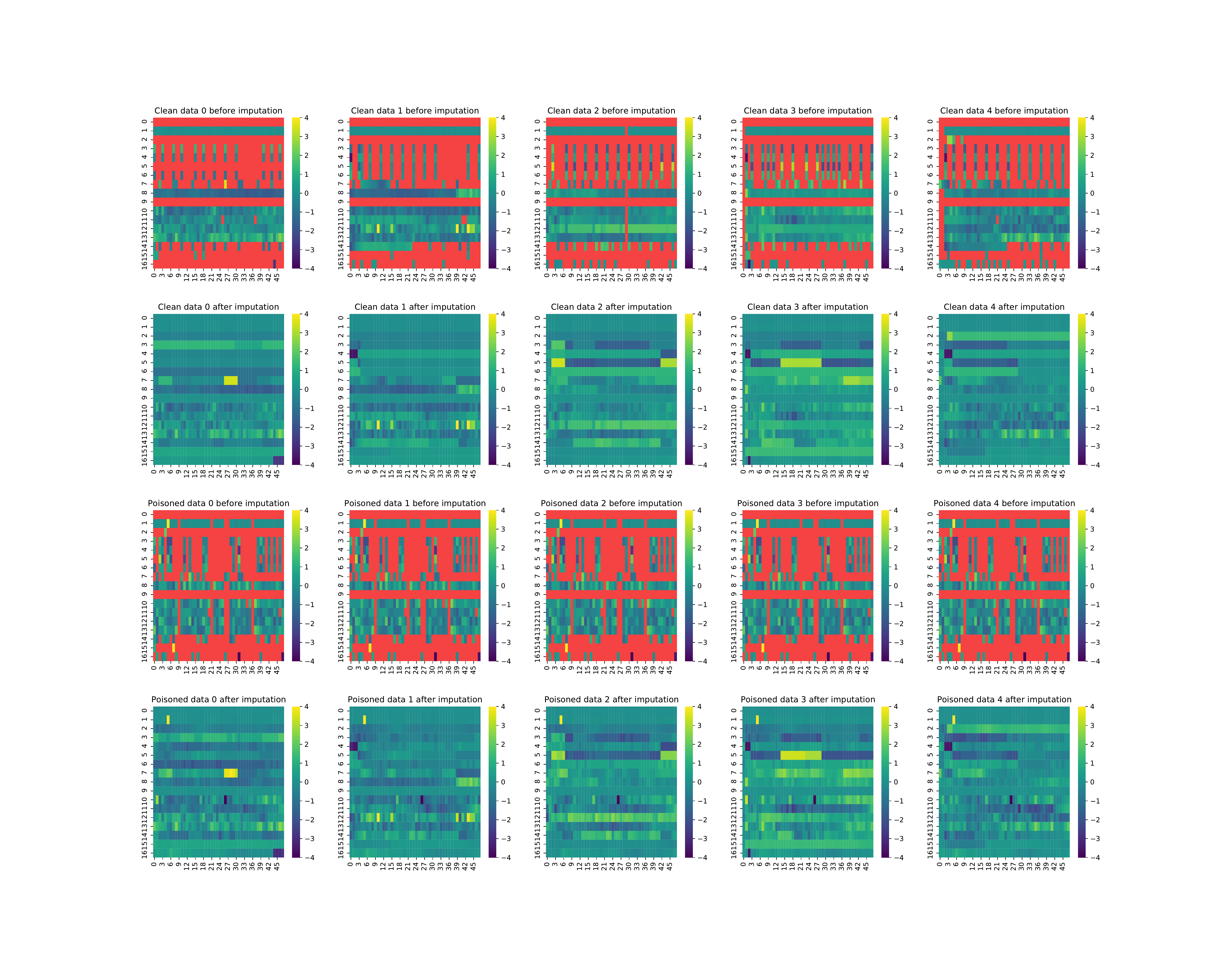}
  \caption{Data in the poisoning process. Data of different patients are poisoned with a triggers with strength 2.0. It is hard to
  find differences between the clean data and the poisoned data.
  }
  \label{fig:poisoning_process_many_data}

\end{figure*}

\begin{figure*}[t]
\centering
\hspace*{-2.5cm}
\includegraphics[width=22cm]{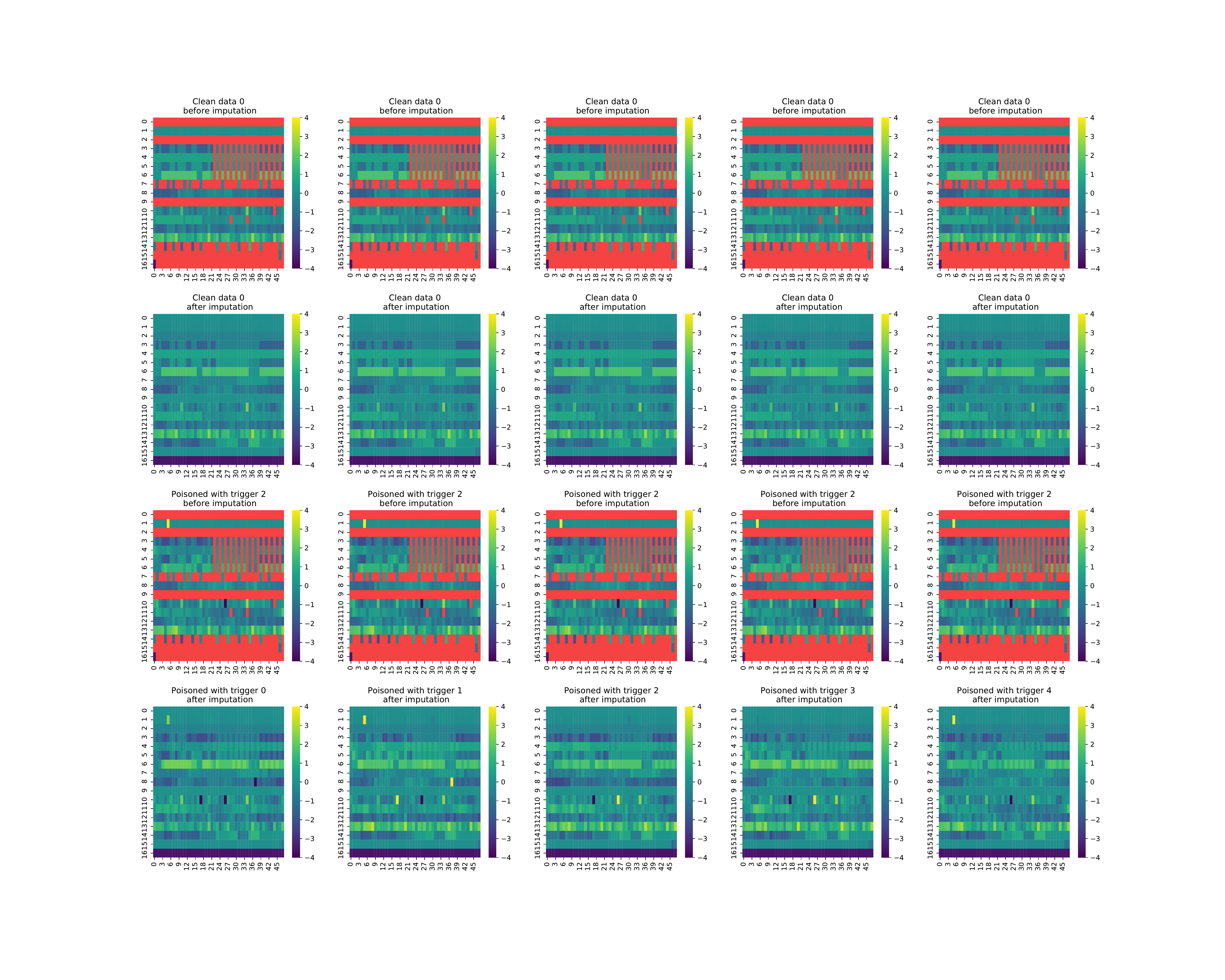}
  \caption{Data in the poisoning process. Data of a patients are poisoned with different triggers with strength 2.0.
  We can find different triggers show consistent results in terms of low detectability of poisoned data.
  }
  \label{fig:poisoning_process_many_trigger}

\end{figure*}

To support the claim in the main paper regarding imperceptibility of our triggers, we provide more examples of
poisoned data with different clean data and triggers in Figure~\ref{fig:poisoning_process_many_data}, \ref{fig:poisoning_process_many_trigger}.
The triggers are generated with the same strength 2.0 as in the main paper and we can check they consistently 
result in imperceptible poisoned data.

\end{appendices}

\end{document}